\pdfoutput=1
\documentclass[10pt,twocolumn,letterpaper]{article}

\usepackage[pagenumbers]{wacv}              % To produce the CAMERA-READY version

\usepackage{booktabs}    % \toprule \midrule \bottomrule \cmidrule
\usepackage{multirow}    % \multirow
\usepackage{graphicx}    % \resizebox
\usepackage{xcolor}      % \textcolor{red}{...}
\usepackage{makecell}
\usepackage{mathtools}   % psmallmatrix (supplementary)

\usepackage{pifont}
\newcommand{\cmark}{\ding{51}}
\newcommand{\xmark}{\ding{55}}

\usepackage{amssymb}
\usepackage[utf8]{inputenc}
\usepackage{textgreek}
\usepackage{bm}

\definecolor{wacvblue}{rgb}{0.21,0.49,0.74}
\usepackage[pagebackref,breaklinks,colorlinks,allcolors=wacvblue]{hyperref}

\def\wacvPaperID{2390} % *** Enter the WACV Paper ID here
\def\confName{WACV}
\def\confYear{2027}

\title{DEFDVO: Deformable Deep Visual Odometry for Equirectangular Images}

\author{
Burak Sevsay \quad
Egor Bondarev \quad
Erkut Akdag\\
AIMS Group, Department of Electrical Engineering, Eindhoven University of Technology\\
{\tt\small \{b.sevsay, e.bondarev, e.akdag\}@tue.nl}
}

\begin{document}
\maketitle
\begin{abstract}
Omnidirectional ($360^\circ$) cameras capture the full surrounding scene in every frame, providing a strong advantage for visual odometry (VO) over pinhole sensors with a limited field of view. However, their equirectangular projection (ERP) introduces latitude-dependent distortion, an artificial seam at the image border, and a sampling bias toward the poles. Existing $360^\circ$ VO methods address these challenges only partially. DEFDVO closes this gap with a patch-based deep VO method that operates directly on ERP imagery, combining spherically initialized deformable feature extraction, end-to-end seam-aware estimation, and spherical-gradient patch selection. The deformable feature extractor combines spherical geometric priors with learned sampling offsets, while seam-aware processing preserves ERP continuity, and spherical-gradient selection reduces polar sampling bias. Trained exclusively on synthetic TartanAirV2 ERP imagery, DEFDVO improves the best prior average absolute trajectory error (ATE) by $4.8\times$ (0.201 m to 0.042 m) on the TartanAirV2 evaluation set defined by 360DVO and achieves the lowest average ATE on both splits of the real-world 360DVO benchmark. Code is available at \url{https://github.com/BurakSevsay/DEFDVO}.
\end{abstract}

\section{Introduction}
\label{sec:intro}

Visual odometry (VO) estimates camera motion from sequential images and is a fundamental component of visual simultaneous localization and mapping (VSLAM). Recent learning-based methods have advanced VO by replacing handcrafted feature matching with learned representations and optimization. However, these methods are designed for conventional perspective cameras with a limited field of view. As a result, they become unreliable under large rotations or abrupt motion, where the overlap between consecutive frames is significantly reduced.

\begin{figure}[t]
\centering
\includegraphics[width=\columnwidth]{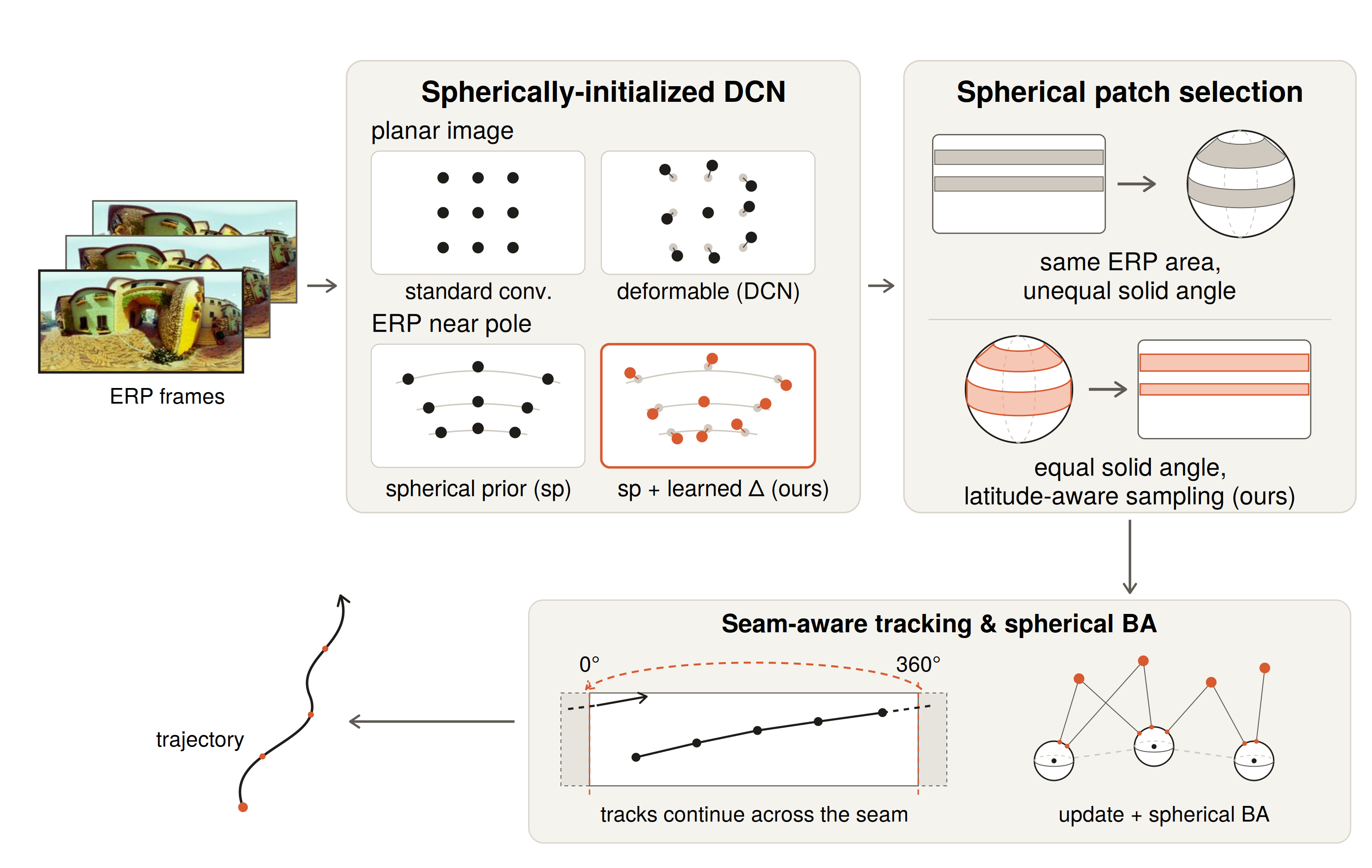}
  \caption{\textbf{DEFDVO overview.}
  Given an equirectangular (ERP) stream, a spherically initialized deformable encoder extracts distortion-aware features, spherical-gradient selection samples informative patches on the viewing sphere, and seam-aware tracking with spherical bundle adjustment estimates the camera trajectory.}
  \label{fig:overview}
\end{figure}

Omnidirectional ($360^\circ$) cameras overcome this limited field of view by capturing the full viewing sphere in every frame, thereby preserving visual overlap even under large rotations. However, extending VO from perspective to $360^\circ$ imagery introduces a different challenge: conventional image-space operations do not conform to the underlying spherical geometry. The equirectangular projection (ERP) provides a convenient rectangular representation of $360^\circ$ imagery, but it introduces three key geometric challenges. First, distortion increases with latitude, while standard convolutions use a fixed rectangular sampling grid. Second, ERP is periodic in azimuth, but conventional methods treat its left and right boundaries as disconnected, breaking tracks that cross the seam. Third, a pixel at latitude $\phi$ subtends a solid angle proportional to $\cos\phi$, causing uniform image-space sampling to over-represent polar regions.

Recent learning-based omnidirectional VO methods address these challenges only partially. CubeDVO~\cite{ando2026cubedvo} avoids ERP distortion through cubemap projection but introduces face boundaries and additional resampling. 360DVO~\cite{guo2026_360dvo} operates directly on ERP imagery with spherical feature extraction and omnidirectional differentiable bundle adjustment. However, its feature sampling  is fixed by ERP geometry and does not adapt to local image content. Additionally, azimuth periodicity and non-uniform spherical sampling are not explicitly addressed in correspondence estimation and patch selection. These limitations motivate a unified approach that preserves ERP geometry throughout the patch-based VO pipeline while ensuring the flexibility of learned feature sampling.

We propose \textbf{DEFDVO}, a monocular patch-based visual odometry framework that builds on DPVO~\cite{teed2023dpvo} and operates directly on equirectangular imagery (Fig.~\ref{fig:overview}). The proposed method addresses ERP distortion, azimuth periodicity, and non-uniform spherical sampling through spherically initialized deformable feature extraction, end-to-end seam-aware estimation, and spherical-gradient patch selection. The main contributions are summarized as follows:

\begin{itemize}
\item Spherically initialized deformable feature extraction: a DCNv2-based~\cite{zhu2019dcnv2} encoder whose sampling offsets combine a frozen analytic SphereNet prior with learned, latitude-conditioned residuals, yielding distortion-aware yet content-adaptive features (Sec.~\ref{sec:encoder}).

\item End-to-end seam-aware estimation: circular padding in the encoder, circularly indexed correlation lookup, azimuth-modular bundle-adjustment residuals, and a wrapped training loss, preserving ERP periodicity throughout the pipeline (Sec.~\ref{sec:seam}--\ref{sec:ba}).

\item Spherical-gradient patch selection: latitude-aware sampling that reduces the $1/\cos\phi$ bias of planar ERP proposals while maintaining high-gradient regions (Sec.~\ref{sec:patchsel}).
\end{itemize}
\section{Related Work}
\label{sec:related_work}

\begin{figure*}[t]
\centering
\includegraphics[width=\linewidth,]{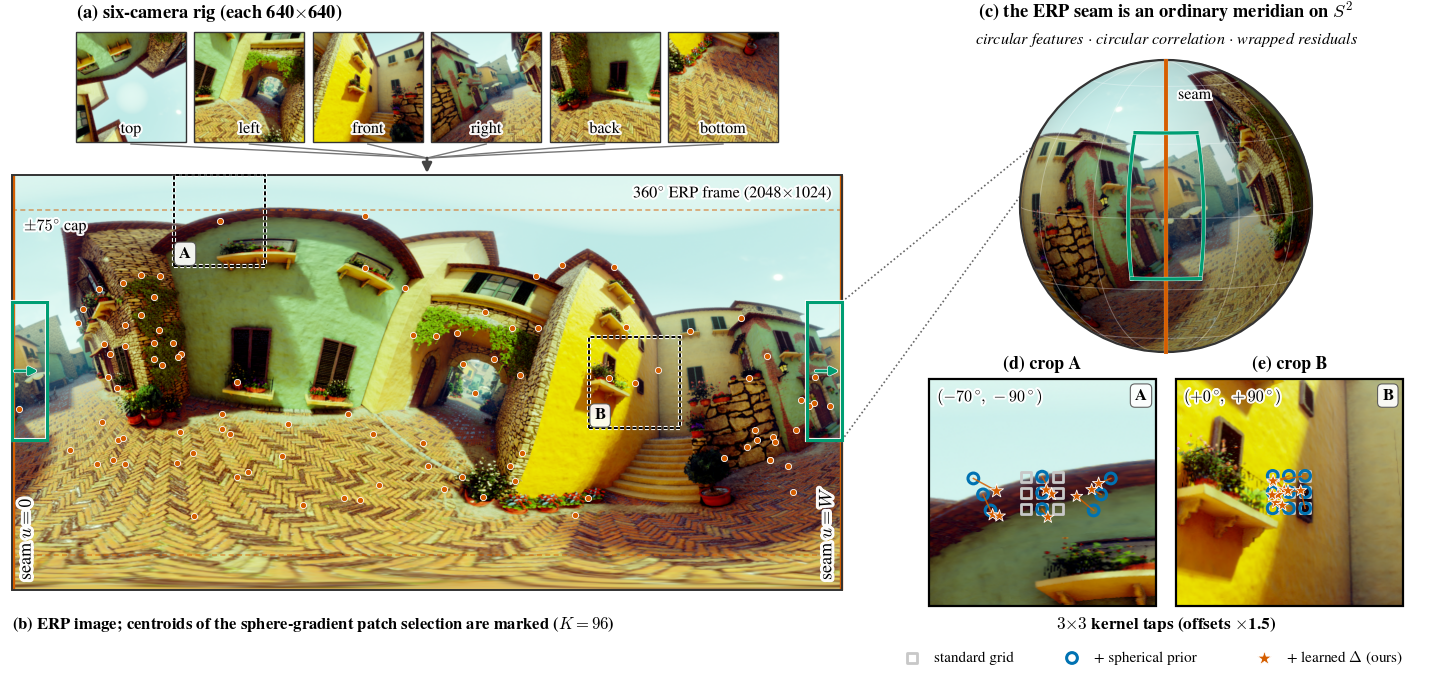}
\caption{Overview of DEFDVO on TartanAirV2 ERP imagery. (a) The six $640\times640$ rig views are combined into a single $2048\times1024$ equirectangular (ERP) frame. (b) ERP image with spherical-gradient patch centroids and the $\pm75^\circ$ latitude cap (Sec.~\ref{sec:patchsel}). The green half-boxes represent continuity on the ERP seam. (c) On the sphere, the seam is an ordinary meridian, and the pipeline treats it consistently end-to-end through circular feature extraction and correlation and azimuth-wrapped bundle-adjustment residuals. (d,e) $3\times3$ deformable-kernel taps at the two crops marked A/B in (b): the regular sampling grid (gray squares), analytic spherical prior (blue circles), and prior plus the learned residual of the trained encoder (orange stars).}
\label{fig:method_overview}
\end{figure*}

\subsection{Learning-Based Visual Odometry}
\noindent\textit{Perspective methods.}
DROID-SLAM~\cite{teed2021droid} introduces recurrent correspondence refinement coupled with differentiable bundle adjustment, and DPVO~\cite{teed2023dpvo} sparsifies this formulation into a patch-to-frame graph that jointly refines correspondences, camera poses, and patch depths. DPV-SLAM~\cite{lipson2024dpvslam} extends the patch-based framework with loop closure and global optimization. More recently, MASt3R-SLAM~\cite{murai2025mast3rslam} builds dense monocular SLAM around learned two-view 3D reconstruction and matching priors, and VGGT-SLAM~\cite{maggio2025vggtslam} incrementally aligns submaps produced by the feed-forward VGGT reconstruction model under global projective optimization. DEFDVO follows the sparse recurrent design of DPVO for its efficiency but adapts its planar image-space operations to equirectangular geometry.

\noindent\textit{Omnidirectional methods.}
A complementary line of work adapts feature extraction, camera models, or photometric objectives to wide field-of-view imagery. Representative examples include spherical binary features~\cite{zhao2015sphorb}, ERP support in OpenVSLAM~\cite{sumikura2019openvslam}, omnidirectional DSO~\cite{matsuki2018omnidso}, 360VO~\cite{huang2022_360vo}, and multi-fisheye SLAM~\cite{won2020omnislam}. The most closely related learning-based approaches to DEFDVO are 360DVO~\cite{guo2026_360dvo} and CubeDVO~\cite{ando2026cubedvo}. 360DVO extends DPVO-style sparse patch tracking to ERP imagery using a SphereNet-based, distortion-aware feature extractor and omnidirectional differentiable bundle adjustment. CubeDVO, instead, projects ERP images onto cubemap faces for feature processing and performs geometric optimization on the unit sphere, trading ERP distortion for face-boundary discontinuities and additional resampling. In contrast, DEFDVO operates directly on the native ERP representation and targets all three issues identified in Sec.~\ref{sec:intro}.

\subsection{Distortion-Aware and Deformable Feature Extraction}
ERP distortion varies with latitude, whereas standard convolutions apply a fixed rectangular sampling grid. SphereNet~\cite{coors2018spherenet} adapts kernel sampling analytically according to spherical geometry. Related methods use distortion-aware filters~\cite{tateno2018distortion}, analytically computed ERP offsets~\cite{fernandez2020corners}, or transformed pinhole kernels~\cite{su2019ktn}. Deformable convolution provides a complementary, purely data-driven alternative that learns spatial sampling offsets~\cite{dai2017dcn}. DCNv2 adds learned per-sample modulation~\cite{zhu2019dcnv2}, and DCNv3/v4 scale the operator and optimize its implementation~\cite{wang2023internimage,xiong2024dcnv4}. These two families have largely been studied in isolation, and only fixed analytic sampling has been employed~\cite{guo2026_360dvo} within sparse omnidirectional VO. The DEFDVO encoder combines them: an analytic spherical prior provides a geometry-aware initialization, while learned latitude-conditioned residual offsets allow for task- and content-dependent adaptation.

\subsection{Seam-Aware Panoramic Processing}
Optical-flow methods for $360^\circ$ imagery address the ERP seam with increasing directness. OmniFlowNet~\cite{artizzu2020omniflownet} adapts a pinhole flow network to ERP distortion with lookup tables, and SLOF~\cite{bhandari2022slof} improves robustness through rotation augmentation and Siamese consistency. PanoFlow~\cite{shi2023panoflow} goes further by defining \emph{cyclic} flow displacements modulo the azimuth circle, showing that seam-aware estimation substantially reduces effective displacement magnitudes. In dense $360^\circ$ perception, circular and cube padding are standard seam treatments~\cite{cheng2018cube,wang2020bifuse,jiang2021unifuse,reyarea2022_360monodepth}. These methods make the features or the output representation seam-aware. However, in omnidirectional VO, azimuth periodicity is typically not enforced consistently across feature extraction, correlation lookup, optimization residuals, and training losses. DEFDVO preserves this periodicity end to end in a patch-based VO pipeline.

\subsection{Patch Selection on Spherical Imagery}
\label{subsect:patch_select}
The accuracy of sparse VO also depends on where patches are placed. DPVO~\cite{teed2023dpvo} samples patch centroids uniformly at random or according to image-gradient magnitude, in both cases on the planar image grid. Classical omnidirectional pipelines, instead, adapt interest-point detection to spherical geometry, e.g., through binary features defined on a geodesic grid~\cite{zhao2015sphorb,guan2017brisks}, and through tangent-image representations that mitigate distortion by locally re-projecting the sphere before detection~\cite{eder2020tangent}. Learned ERP-based VO, however, relies on planar patch selection, which ignores the $\cos\phi$ solid-angle weighting of the ERP grid. Our spherical-gradient selection closes this gap by weighting image gradients by per-pixel solid angle and sampling patch candidates in proportion to solid angle rather than pixel area.

\newcommand{\wrap}{\operatorname{wrap}}
\newcommand{\atantwo}{\operatorname{atan2}}
\newcommand{\SE}{\mathrm{SE}(3)}
\newcommand{\Sim}{\mathrm{Sim}(3)}

\section{Method}
\label{sec:method}

DEFDVO builds on the sparse recurrent formulation of DPVO~\cite{teed2023dpvo}, which iteratively refines patch correspondences and jointly optimizes camera poses and patch geometry through differentiable bundle adjustment (BA). DEFDVO preserves this optimization framework while adapting its image-space operations to equirectangular geometry through three components: spherically initialized deformable feature extraction, seam-aware correspondence and optimization, and spherical-gradient patch selection.

\subsection{Overview and notation}
Figure~\ref{fig:method_overview} illustrates the geometry-aware components of DEFDVO on ERP imagery. An ERP image has a width $W$ and a height $H=W/2$, with pixel coordinates $(u,v)$. Camera poses $\mathbf{T}_i \in \SE$ are world-to-camera rigid-body transformations in the special Euclidean group, and the relative transformation from frame $i$ to frame $j$ is $\mathbf{G}_{ij} = \mathbf{T}_j \mathbf{T}_i^{-1} = (\mathbf{R}_{ij}, \mathbf{t}_{ij})$, with rotation $\mathbf{R}_{ij}$ and translation $\mathbf{t}_{ij}$. Each patch $k$, hosted in frame $i$, is a $3{\times}3$ pixel grid sharing a common inverse range $d_k$. Following DPVO~\cite{teed2023dpvo}, the patch graph connects each hosted patch to target frames within the active optimization window.

\subsection{Equirectangular Camera Model with Inverse Range}
\label{sec:model}
DPVO assumes perspective projection and parameterizes scene structure using inverse $z$-depth, which is not suitable for full-sphere ERP imagery. DEFDVO instead adopts a spherical camera model and represents scene structure by inverse Euclidean range, providing a consistent parameterization over the full viewing sphere.

For an ERP image, pixel coordinates $(u,v)$ map linearly to azimuth
$\theta$ and elevation $\phi$ (positive downwards) as
$\theta=(u-c_x)/f_x$ and $\phi=(v-c_y)/f_y$, where $f_x=W/(2\pi)$, $f_y=H/\pi$, $c_x=W/2$, and $c_y=H/2$. Using camera axes $x$-right, $y$-down, and $z$-forward, the corresponding unit bearing vector is 
\begin{equation}
\mathbf{n}(\theta,\phi)=
\big(
\sin\theta\cos\phi,\;
\sin\phi,\;
\cos\theta\cos\phi
\big)^{\!\top}.
\label{eq:bearing}
\end{equation}

Unlike inverse $z$-depth, inverse \emph{Euclidean range} remains positive and well defined over the full viewing sphere. Each patch is therefore assigned as $d=1/\lVert\mathbf{P}\rVert$, where $\mathbf{P}$ is the patch center in its host-camera frame. The $z$-depth vanishes for viewing directions perpendicular to the optical axis and changes sign behind the camera, whereas the Euclidean range remains positive and well defined over the full viewing sphere. Each patch center is stored as a homogeneous point $\tilde{\mathbf{X}}=(\mathbf{n}^{\top},d)^{\top}$. This preserves the point representation used by DPVO while replacing perspective depth with spherical range. 

For patch $k$ hosted in frame $i$, the corresponding scale-free
point in target frame $j$ is
\begin{equation}
\mathbf{x}_{kj}
=
\mathbf{R}_{ij}\mathbf{n}_k
+
d_k\mathbf{t}_{ij}
=
(X,Y,Z)^\top .
\label{eq:transform}
\end{equation}
Since ERP projection depends only on viewing direction,
$\mathbf{x}_{kj}$ can be projected directly. Let
$\rho=\sqrt{X^2+Z^2}$. The target-frame spherical coordinates
and ERP projection are
\begin{equation}
\begin{aligned}
\theta' &= \atantwo(X,Z),\\
\phi'   &= \atantwo(Y,\rho),\\
u'_{kj} &= (f_x\theta' + c_x)\bmod W,\\
v'_{kj} &= f_y\phi' + c_y .
\end{aligned}
\label{eq:proj}
\end{equation}
The resulting reprojection is denoted by
$\mathbf{u}'_{kj}
=\Pi(\mathbf{x}_{kj})
=(u'_{kj},v'_{kj})^\top$.
The modulo operation maps azimuth to $[0,W)$, while seam-crossing
operations are handled explicitly in Sec.~\ref{sec:seam}.

\subsection{Seam-Aware Residuals and Circular Correlation}
\label{sec:seam}

The left and right ERP boundaries represent adjacent viewing directions, whereas planar image operations treat them as disconnected, breaking tracks that cross the seam. Since azimuth is periodic with a period of $W=2\pi f_x$ pixels, horizontal differences are computed on the circle:

\begin{equation}
\wrap(\delta)=
\big((\delta+\tfrac{W}{2})\bmod W\big)
-\tfrac{W}{2},
\label{eq:wrap}
\end{equation}
where $\delta$ denotes a horizontal pixel difference and $\wrap(\delta)\in[-W/2,W/2)$. Given the update operator's predicted target $\mathbf u^\star_{kj}=(u^\star_{kj},v^\star_{kj})^\top$, the seam-aware residual is
\begin{equation}
\mathbf r_{kj}=\begin{pmatrix}\wrap(u^\star_{kj}-u'_{kj})\\ v^\star_{kj}-v'_{kj}\end{pmatrix}.
\label{eq:residual}
\end{equation}
The same wrapping is used for horizontal flow-supervision errors during training.

Residual wrapping alone is insufficient because correspondence lookup should also remain continuous across the seam. DEFDVO samples a $7{\times}7$ correlation window around each reprojection at two pyramid levels using bilinear interpolation. Horizontal coordinates are indexed modulo the feature-map width during correlation and patch extraction, while standard vertical bounds checks are applied. Together with circular padding in the encoder (Sec.~\ref{sec:encoder}), this preserves correspondence continuity across the ERP seam.

\subsection{Differentiable Bundle Adjustment with ERP Reprojection}
\label{sec:ba}
DEFDVO builds on DPVO's differentiable BA but replaces planar reprojection and Jacobians with the spherical model in Sec.~\ref{sec:model} and the wrapped residuals in Eq.~\eqref{eq:residual}. Given confidence weights $\mathbf{w}_{kj}\in(0,1)^2$, BA jointly optimizes the window poses and patch inverse ranges by minimizing 

\begin{equation}
E=
\sum_{(k,j)}
\lVert\mathbf r_{kj}\rVert^2_{\operatorname{diag}(\mathbf w_{kj})}.
\label{eq:energy}
\end{equation}

Each damped Gauss--Newton iteration linearizes the residuals with
respect to pose and inverse range. For the transformed point
$\mathbf{x}_{kj}=(X,Y,Z)^\top$, define
$s=\|\mathbf{x}_{kj}\|$ and $\rho=\sqrt{X^2+Z^2}$.
The ERP projection Jacobian is
\begin{equation}
\label{eq:jpi}
\mathbf{J}_\Pi
=
\frac{\partial(u',v')}{\partial\mathbf{x}}
=
\begin{pmatrix}
\dfrac{f_x Z}{\rho^2} & 0 & -\dfrac{f_x X}{\rho^2} \\[2ex]
-\dfrac{f_y X Y}{\rho s^2} &
\dfrac{f_y \rho}{s^2} &
-\dfrac{f_y Z Y}{\rho s^2}
\end{pmatrix}.
\end{equation}
The row norms of $\mathbf{J}_{\Pi}$ are
$f_x/(s\cos\phi')$ and $f_y/s$, where $\phi'$ is the elevation of the reprojection. Since $W=2H$ implies $f_x=f_y$, horizontal sensitivity increases as $1/\cos\phi'$ toward the poles. Near-polar measurements can, therefore,
disproportionately influence the normal equations, motivating the
latitude-aware patch selection in Sec.~\ref{sec:patchsel}. The remaining pose and inverse-range Jacobians follow from $\mathbf{J}_{\Pi}$ through the chain rule and are provided in the supplementary material.

As in DPVO, each Gauss–Newton step eliminates inverse ranges through a Schur complement, anchors the optimization with fixed reference poses, updates poses multiplicatively, and inverse ranges additively. Two differentiable iterations are used during training, with an equivalent CUDA solver at inference. Invalid patch–frame pairs are masked using range validity and vertical image bounds, while learned confidence weights provide robustness to outliers.

\subsection{Spherically Initialized Deformable Feature Extraction}
\label{sec:encoder}
Standard convolutions use a fixed rectangular sampling grid and, therefore, do not account for latitude-dependent ERP distortion. Analytic spherical convolution adapts sampling to known projection geometry but remains independent of image content, whereas deformable convolution learns content-adaptive offsets without an explicit spherical prior. DEFDVO combines these complementary properties using modulated deformable convolution~\cite{zhu2019dcnv2}. As in DPVO, the encoders produce 128-dimensional matching features and 384-dimensional context features at $1/4$ resolution.

For each kernel tap $\kappa$, DEFDVO augments the regular sampling
location with an analytic spherical offset and a learned residual:
\begin{equation}
\mathbf{q}_\kappa(\mathbf{p})
=
\mathbf{p}+\mathbf{p}_\kappa
+\Delta^{\mathrm{sph}}_\kappa(v)
+\Delta^{\mathrm{net}}_\kappa(\mathbf{p}),
\label{eq:dcn}
\end{equation}
where $\mathbf{p}$ is the current feature location,
$\mathbf{p}_\kappa\in\{-1,0,1\}^2$ is the regular $3{\times}3$
kernel offset, $\Delta^{\mathrm{sph}}_\kappa(v)$ is the fixed
geometry-derived offset, and
$\Delta^{\mathrm{net}}_\kappa(\mathbf{p})$ is the learned residual.

The analytic offset $\Delta^{\mathrm{sph}}_\kappa(v)$ follows the SphereNet sampling geometry~\cite{coors2018spherenet}, which maps a regular tangent-plane kernel to ERP coordinates according to latitude. The resulting pattern coincides with the regular grid at the equator and expands horizontally approximately as $1/\cos\phi$ toward the poles. Since the offset depends only on image row, it is precomputed and kept fixed during training.

The learned residual $\Delta^{\mathrm{net}}_\kappa(\mathbf p)$ provides content-adaptive refinement of the geometric prior. A lightweight predictor $h_\theta$ takes local features together with $\sin\phi$ and $\cos\phi$ latitude maps and predicts residual offsets and modulation logits $\ell_\kappa$. The corresponding gate $m_\kappa=2\sigma(\ell_\kappa)\in(0,2)$ modulates each sampling location. Since $h_\theta$ is zero-initialized, training starts with $\Delta^{\mathrm{net}}=0$ and $m_\kappa=1$, making the layer equivalent to analytic spherical convolution at initialization. Circular padding in azimuth allows deformable samples to cross the ERP seam continuously.

\subsection{Patch Selection on the Sphere}
\label{sec:patchsel}

Uniform sampling on the ERP grid over-represents polar regions because the solid angle represented by each pixel scales with $\cos\phi$. Moreover, ERP reprojection becomes increasingly ill-conditioned toward the poles (Sec.~\ref{sec:ba}). DEFDVO addresses both issues through spherical-area sampling followed by latitude-aware gradient ranking. Candidate latitudes are first sampled according to
\begin{equation}
p(\phi)\propto\cos\phi,
\qquad |\phi|\le75^\circ,
\label{eq:area_sampling}
\end{equation}
% with azimuth sampled uniformly. Candidates are then oversampled by $3\times$ and ranked by
with azimuth sampled uniformly. The $75^\circ$ cap is a safeguard against near-polar patches rather than a tuned threshold, as detailed in the supplementary material. Candidates are then oversampled by $3\times$ and ranked by
\begin{equation}
s(\mathbf p)=g(\mathbf p)\cos\phi(\mathbf p),
\label{eq:patch_score}
\end{equation}
where $g(\mathbf p)$ is the feature-gradient magnitude. The two latitude factors serve different purposes: spherical-area sampling corrects candidate density on the sphere, whereas the ranking weight suppresses poorly conditioned high-latitude measurements. The same strategy is used during training and inference.

\subsection{Training}
\label{sec:training}
DEFDVO follows DPVO~\cite{teed2023dpvo} for pose supervision while adapting correspondence supervision to ERP periodicity. The training objective is
\begin{equation}
\mathcal{L}=0.1\,\mathcal{L}_{\mathrm{flow}}+10\,\mathcal{L}_{\mathrm{pose}}
\end{equation}
The pose loss is computed after $\Sim$ alignment with the ground-truth trajectory.

Standard endpoint error is discontinuous at the ERP seam, where adjacent spherical locations can be separated by nearly the full image width. DEFDVO applies the wrapped displacement in Eq.~\eqref{eq:wrap} to horizontal flow errors.

\begin{equation}
\mathcal{L}_{\mathrm{flow}} =
\operatorname*{mean}_{(k,j)\,\mathrm{valid}}\;
\min_{\mathbf{p} \in \mathrm{patch}}
\sqrt{
\wrap_{W\!/4}(\Delta u)^2
+
\Delta v^2
+
\epsilon
},
\label{eq:flowloss}
\end{equation}

where $(\Delta u,\Delta v)$ denotes the difference between the predicted and ground-truth flow at patch pixel $\mathbf p$, and $\wrap_{W/4}$ applies Eq.~\eqref{eq:wrap} at the $1/4$-resolution feature grid. The minimum is taken over the $3{\times}3$ patch support. Invalid patch-frame pairs are excluded from the loss.

\section{Experiments}
\label{sec:experiments}

\begin{table*}[t]
\centering
\caption{ATE/RPE(t)/RPE(r) on TartanAirV2 Set~1 (360DVO evaluation scenes). Input: Pin.\ = pinhole, ERP = equirectangular. Best ATE per sequence in \textbf{bold}, second-best \underline{underlined}.}
\label{tab:set1}
\setlength{\tabcolsep}{3pt}
\resizebox{\textwidth}{!}{%
\begin{tabular}{llccccccccc}
\toprule
\multirow{2}{*}{Methods} & \multirow{2}{*}{Input} & \multicolumn{2}{c}{CountryHouse} & \multicolumn{2}{c}{House}
 & \multicolumn{2}{c}{OldTownNight} & \multicolumn{2}{c}{VictorianStreet}
 & \multirow{2}{*}{Avg.} \\
\cmidrule(lr){3-4}\cmidrule(lr){5-6}\cmidrule(lr){7-8}\cmidrule(lr){9-10}
 & & Easy & Hard & Easy & Hard & Easy & Hard & Easy & Hard & \\
\midrule
DROID-SLAM~\cite{teed2021droid} & Pin. & \textbf{0.002}/0.000/0.013 & \textbf{0.003}/0.001/0.015 & \textbf{0.015}/0.001/0.021 & 1.893/0.125/0.316 & 3.951/0.025/0.057 & 3.422/0.139/2.004 & \textbf{0.009}/0.001/0.027 & \textbf{0.017}/0.002/0.015 & 1.164/0.037/0.308 \\
DPVO~\cite{teed2023dpvo} & Pin. & 0.012/0.001/0.015 & 0.011/0.001/0.021 & \underline{0.030}/0.001/0.022 & 2.000/0.131/0.884 & 2.893/0.020/0.105 & 8.960/0.289/3.378 & 0.052/0.002/0.034 & 0.090/0.003/0.020 & 1.756/0.056/0.560 \\
MASt3R-SLAM~\cite{murai2025mast3rslam} & Pin. & 0.041/0.010/0.256 & 0.043/0.017/0.425 & 0.231/0.014/0.274 & 0.476/0.068/2.047 & 8.659/0.193/2.083 & 10.465/0.934/6.869 & 0.397/0.040/0.813 & 0.703/0.065/1.622 & 2.627/0.168/1.798 \\
VGGT-SLAM~\cite{maggio2025vggtslam} & Pin. & 0.145/0.048/1.181 & 0.165/0.070/1.136 & 0.594/0.157/1.547 & 0.333/0.100/2.139 & 1.375/0.345/1.057 & 7.208/1.845/10.315 & 0.729/0.163/1.520 & 0.683/0.177/1.306 & 1.404/0.363/2.525 \\
\midrule
OpenVSLAM~\cite{sumikura2019openvslam} & ERP & 0.783/0.266/3.223 & 0.621/0.373/57.328 & 2.025/1.096/43.676 & 1.169/0.607/10.045 & 3.848/0.651/16.078 & 3.204/0.493/8.473 & 3.371/2.223/8.151 & 5.347/3.110/36.991 & 2.546/1.102/22.996 \\
360DVO (default)~\cite{guo2026_360dvo} & ERP & 0.012/0.001/0.008 & \underline{0.005}/0.001/0.015 & 0.038/0.001/0.012 & \textbf{0.042}/0.003/0.024 & \underline{0.145}/0.003/0.008 & 1.260/0.029/0.035 & 0.066/0.002/0.008 & 0.046/0.002/0.011 & 0.202/0.005/0.015 \\
360DVO (fast) & ERP & 0.024/0.002/0.021 & 0.015/0.002/0.033 & 0.053/0.002/0.028 & 0.088/0.006/0.043 & 0.207/0.011/0.094 & \underline{1.097}/0.028/0.055 & 0.080/0.014/0.045 & 0.046/0.003/0.024 & \underline{0.201}/0.008/0.043 \\
DEFDVO (ours) & ERP & \underline{0.011}/0.001/0.015 & \underline{0.005}/0.001/0.016 & 0.032/0.002/0.020 & \underline{0.049}/0.004/0.027 & \textbf{0.055}/0.007/0.024 & \textbf{0.116}/0.013/0.049 & \underline{0.040}/0.018/0.044 & \underline{0.031}/0.003/0.015 & \textbf{0.042}/0.006/0.026 \\

\bottomrule
\end{tabular}}
\end{table*}

\newlength{\scenecol}
\newcolumntype{C}{>{\centering\arraybackslash}p{\scenecol}}

\begin{table*}[t]
\centering
\caption{ATE on TartanAirV2 Set~2 (CubeDVO evaluation scenes). Input: Pin.\ = pinhole, ERP = equirectangular, Cube.\ = cubemap. ``--'' denotes tracking failure. Best ATE per sequence in \textbf{bold}, second-best \underline{underlined}.}
\label{tab:set2}
\setlength{\tabcolsep}{3pt}
\settowidth{\scenecol}{DesertGasStation}% widest scene name sets the group width
\setlength{\scenecol}{\dimexpr0.5\scenecol-\tabcolsep+1pt\relax}% each pair = name width + 2pt slack
\resizebox{\textwidth}{!}{%
\begin{tabular}{@{}lc*{12}{C}c@{}}
\toprule
\multirow{2}{*}{Methods} & \multirow{2}{*}{Input} & \multicolumn{2}{c}{Antiquity3D} & \multicolumn{2}{c}{DesertGasStation} & \multicolumn{2}{c}{HongKong} & \multicolumn{2}{c}{HQWesternSaloon} & \multicolumn{2}{c}{IndustrialHangar} & \multicolumn{2}{c}{OldTownNight} & \multirow{2}{*}{Avg.} \\
\cmidrule(lr){3-4}\cmidrule(lr){5-6}\cmidrule(lr){7-8}\cmidrule(lr){9-10}\cmidrule(lr){11-12}\cmidrule(lr){13-14}
 & & Easy & Hard & Easy & Hard & Easy & Hard & Easy & Hard & Easy & Hard & Easy & Hard & \\
\midrule
DROID-SLAM~\cite{teed2021droid} & Pin. & 12.460 & 4.267 & 0.103 & -- & \textbf{0.034} & 0.164 & \textbf{0.008} & 2.446 & -- & -- & 3.951 & \underline{0.053} & -- \\
DPVO~\cite{teed2023dpvo} & Pin. & \underline{0.119} & 2.892 & 0.128 & 9.236 & 0.065 & 1.558 & \underline{0.033} & 3.060 & 0.634 & 7.002 & 2.893 & 0.200 & 2.318 \\
MASt3R-SLAM~\cite{murai2025mast3rslam} & Pin. & 17.682 & 17.491 & 9.196 & 9.281 & -- & 0.491 & 0.205 & 6.035 & 19.972 & 20.843 & 8.659 & 0.663 & -- \\
VGGT-SLAM~\cite{maggio2025vggtslam} & Pin. & 1.825 & 1.874 & 1.999 & 2.324 & 1.183 & 2.990 & 1.383 & 1.893 & 6.854 & 7.984 & 1.375 & 3.154 & 2.903 \\
\midrule
OpenVSLAM~\cite{sumikura2019openvslam} & ERP & 1.621 & 1.273 & 7.715 & 9.446 & 1.285 & 0.115 & 4.844 & 5.684 & 15.293 & 4.822 & 3.848 & 0.862 & 4.734 \\
360DVO (default)~\cite{guo2026_360dvo} & ERP & 0.187 & \underline{0.224} & \textbf{0.062} & \textbf{0.064} & \underline{0.062} & \underline{0.106} & 0.052 & \textbf{0.043} & \textbf{0.155} & 0.427 & \underline{0.145} & 0.179 & \underline{0.142} \\
360DVO (fast) & ERP & 0.253 & 0.403 & 0.107 & 0.121 & 0.172 & 0.172 & 0.115 & 0.138 & 0.760 & \underline{0.423} & 0.207 & 2.180 & 0.421 \\
CubeDVO~\cite{ando2026cubedvo} & Cube. & 3.610 & 2.940 & 1.060 & 0.490 & 0.290 & 0.230 & 0.630 & 0.220 & 0.830 & 0.530 & 0.790 & 0.560 & 1.020 \\
DEFDVO (ours) & ERP & \textbf{0.072} & \textbf{0.104} & \underline{0.073} & \underline{0.096} & 0.084 & \textbf{0.074} & 0.047 & \underline{0.047} & \underline{0.189} & \textbf{0.167} & \textbf{0.055} & \textbf{0.052} & \textbf{0.088} \\

\bottomrule
\end{tabular}}
\end{table*}

The evaluation compares DEFDVO on synthetic and real-world $360^\circ$ imagery with the learning-based omnidirectional VO methods 360DVO~\cite{guo2026_360dvo} and CubeDVO~\cite{ando2026cubedvo}, the conventional omnidirectional VO method OpenVSLAM~\cite{sumikura2019openvslam}, and pinhole baselines run on perspective views.

\subsection{Datasets}
\label{sec:datasets}
\emph{TartanAirV2.} The training set contains $2048{\times}1024$ ERP images from TartanAirV2~\cite{wang2020tartanair}, providing ground-truth camera poses and depth information across 65 diverse synthetic scenes, each with easy and hard trajectories. 

360DVO and CubeDVO evaluate on different TartanAirV2 scene subsets, sharing only the OldTownNight easy trajectory. To enable direct comparison under both protocols, the two subsets are evaluated separately: \emph{Set~1} corresponds to the 360DVO evaluation split, whereas \emph{Set~2} corresponds to the CubeDVO split. DEFDVO follows the 360DVO training protocol, using the easy P000 and hard P000--P001 trajectories from each training scene. All scenes in the union of Set~1 and Set~2 are excluded from training, resulting in approximately 194k training images.

\noindent\emph{360DVO dataset.} The real-world 360DVO dataset~\cite{guo2026_360dvo} serves as the second evaluation benchmark. It contains 20 sequences totaling approximately 20k frames, standardized to $3840{\times}1920$ ERP resolution at 10\,FPS. The dataset is divided into easy and hard subsets, with 10 sequences each, based on trajectory complexity and scene dynamics. All ablations are performed on this benchmark.

\subsection{Evaluation Metrics}
\label{sec:metrics}
The evaluation reports absolute trajectory error (ATE, m) after $\Sim$ alignment, together with one-frame relative pose error in translation, RPE(t) [m/frame], and rotation, RPE(r) [deg/frame]. Lower values indicate better performance. Set~2 reports only ATE because CubeDVO does not publish RPE results, and its implementation is unavailable. Following DPVO~\cite{teed2023dpvo}, each sequence is evaluated over five runs, and the median is reported. Split-level results are averaged over the corresponding sequence medians. A method is considered to have failed on a sequence if tracking fails in at least three of the five runs. No split-level average is reported if the method fails on any sequence.

\subsection{Implementation Details}
\label{sec:impl}
DEFDVO is trained from scratch following the DPVO training strategy~\cite{teed2023dpvo}. 
Unlike DPVO, planar crop and scale augmentation are disabled,  and processing scales are restricted to preserve the ERP relation $f_x = W'/(2\pi)$ for the processed width $W'$. Training is performed for 120k steps at $0.5$ image scale and takes approximately two days on an NVIDIA H100 GPU. At inference, DEFDVO uses 192 patches per frame at an image scale of 0.5. On the 360DVO dataset, this corresponds to a $1920{\times}960$ resolution and yields 7.1\,FPS. All inference and evaluation experiments are conducted on an NVIDIA RTX A5000 GPU.

\begin{table*}[t]
\centering
\caption{ATE/RPE(t)/RPE(r) on the 360DVO real-world dataset
(\textit{Easy} and \textit{Hard}). Input: Pin.\ = pinhole, ERP =
equirectangular. ``--'' denotes failure. Best ATE
per sequence in \textbf{bold}, second-best \underline{underlined}.}
\label{tab:360dvo_dataset}
\setlength{\tabcolsep}{4pt}
\setlength{\aboverulesep}{0pt}
\setlength{\belowrulesep}{0pt}
\renewcommand{\arraystretch}{1.2}
\resizebox{\textwidth}{!}{%
\begin{tabular}{ll|cccccccccc|c}
\toprule
\multirow{2}{*}{Methods} & \multirow{2}{*}{Input} & \multicolumn{10}{c|}{\textit{Easy} Sequences} & \multirow{2}{*}{Avg.} \\
\cmidrule(lr){3-12}
 & & 00 & 01 & 02 & 03 & 04 & 05 & 06 & 07 & 08 & 09 & \\
\midrule
DROID-SLAM~\cite{teed2021droid} & Pin. & 2.28/0.40/0.35 & 0.61/0.13/0.04 & --/--/-- & 22.7/1.08/0.69 & 0.57/0.02/0.03 & 5.27/0.15/0.07 & \underline{0.37}/0.22/0.41 & --/--/-- & 23.7/0.72/1.58 & 2.70/0.04/0.09 & --/--/-- \\
DPVO~\cite{teed2023dpvo} & Pin. & \textbf{1.37}/0.32/0.11 & \textbf{0.13}/0.44/0.11 & 37.5/1.36/0.13 & 6.91/1.33/0.46 & 0.57/0.23/0.03 & 4.89/0.24/0.07 & \textbf{0.35}/0.21/0.52 & 16.4/1.68/0.15 & 7.93/0.88/0.44 & \underline{2.23}/0.41/0.10 & 7.83/0.71/0.21 \\
MASt3R-SLAM~\cite{murai2025mast3rslam} & Pin. & 8.23/0.44/0.40 & 3.99/0.34/0.63 & --/--/-- & --/--/-- & --/--/-- & --/--/-- & --/--/-- & --/--/-- & --/--/-- & 4.25/0.34/0.33 & --/--/-- \\
VGGT-SLAM~\cite{maggio2025vggtslam} & Pin. & 23.14/9.35/35.41 & 26.45/4.36/11.79 & --/--/-- & 36.32/69.75/4.18 & 1.78/10.11/0.89 & --/--/-- & 6.13/3.61/7.73 & 16.04/8.65/3.05 & --/--/-- & 8.56/5.30/7.68 & --/--/-- \\
\midrule
OpenVSLAM~\cite{sumikura2019openvslam} & ERP & 10.59/6.61/22.27 & 2.45/1.42/2.30 & \textbf{17.19}/1.83/0.08 & --/--/-- & \underline{0.52}/0.28/0.04 & 33.21/18.22/19.78 & 23.25/5.01/21.28 & \textbf{0.10}/0.22/0.33 & \textbf{0.25}/0.43/0.12 & 24.91/11.68/0.18 & --/--/-- \\
360DVO (default)~\cite{guo2026_360dvo} & ERP & 2.06/0.04/0.07 & 0.52/0.01/0.01 & 20.59/0.10/0.03 & \underline{3.27}/0.06/0.04 & \textbf{0.46}/0.01/0.04 & \textbf{0.51}/0.01/0.02 & \textbf{0.35}/0.00/0.02 & 2.45/0.03/0.03 & \underline{3.38}/0.03/0.07 & 2.37/0.02/0.07 & \underline{3.60}/0.03/0.04 \\
360DVO (fast) & ERP & 2.16/0.04/0.07 & 0.95/0.04/0.07 & 31.94/0.12/0.04 & 3.32/0.06/0.04 & 6.13/0.06/0.12 & \underline{0.72}/0.02/0.05 & 0.48/0.01/0.02 & 6.07/0.05/0.28 & 5.18/0.04/0.07 & 2.49/0.02/0.07 & 5.94/0.05/0.08 \\
DEFDVO (ours) & ERP & \underline{1.54}/0.03/0.07 & \underline{0.24}/0.03/0.06 & \underline{18.86}/0.09/0.03 & \textbf{2.89}/0.20/0.07 & \textbf{0.46}/0.01/0.03 & 0.80/0.01/0.03 & 0.46/0.01/0.05 & \underline{1.56}/0.02/0.03 & 3.94/0.04/0.07 & \textbf{1.82}/0.02/0.07 & \textbf{3.26}/0.05/0.05 \\
\midrule
\multirow{2}{*}{Methods} & \multirow{2}{*}{Input} & \multicolumn{10}{c|}{\textit{Hard} Sequences} & \multirow{2}{*}{Avg.} \\
\cmidrule(lr){3-12}
 & & 00 & 01 & 02 & 03 & 04 & 05 & 06 & 07 & 08 & 09 & \\
\midrule
DROID-SLAM~\cite{teed2021droid} & Pin. & \textbf{0.20}/0.02/0.04 & 2.93/0.60/2.81 & 2.27/0.20/0.65 & 2.42/0.33/0.43 & 10.2/0.77/0.60 & \textbf{2.72}/0.88/1.41 & \textbf{12.9}/0.61/0.64 & 3.50/0.94/0.66 & 5.48/1.25/0.13 & 29.4/1.27/1.54 & 7.20/0.69/0.89 \\
DPVO~\cite{teed2023dpvo} & Pin. & 0.72/0.71/0.03 & 10.4/0.37/1.96 & 1.34/0.91/0.61 & 0.71/0.38/0.10 & 0.61/0.59/0.06 & 4.03/1.03/1.33 & \underline{19.3}/0.53/0.26 & 1.66/1.19/0.01 & 7.97/1.69/0.12 & 22.5/1.06/1.63 & 6.92/0.85/0.61 \\
MASt3R-SLAM~\cite{murai2025mast3rslam} & Pin. & 4.52/0.80/0.54 & --/--/-- & 25.12/0.85/0.47 & 7.59/0.31/0.24 & 6.44/0.51/0.23 & --/--/-- & --/--/-- & 34.26/1.42/2.65 & --/--/-- & 22.19/2.44/1.76 & --/--/-- \\
VGGT-SLAM~\cite{maggio2025vggtslam} & Pin. & 14.77/23.91/8.34 & 7.08/7.43/18.20 & 5.19/15.38/9.27 & 1.56/6.20/3.99 & 8.41/6.41/2.63 & 5.60/2.54/5.80 & 19.88/15.61/2.31 & 20.83/15.42/6.68 & --/--/-- & 7.52/40.16/13.16 & --/--/-- \\
\midrule
OpenVSLAM~\cite{sumikura2019openvslam} & ERP & 23.80/3.79/0.30 & 14.29/4.50/37.97 & 25.28/5.95/4.67 & 7.85/1.47/39.74 & 5.61/2.50/18.83 & 10.29/1.02/5.36 & --/--/-- & 34.11/2.67/47.11 & 8.82/1.51/0.33 & --/--/-- & --/--/-- \\
360DVO (default)~\cite{guo2026_360dvo} & ERP & 0.85/0.01/0.01 & 10.43/0.08/2.03 & 5.84/0.23/0.10 & 4.64/0.19/2.91 & \textbf{0.12}/0.00/0.01 & 10.30/0.85/5.35 & 22.91/0.25/0.12 & \underline{1.50}/0.08/0.04 & 3.45/0.06/0.03 & \textbf{0.56}/0.13/0.13 & 6.06/0.19/1.07 \\
360DVO (fast) & ERP & 0.53/0.01/0.01 & \textbf{0.80}/0.04/1.30 & \textbf{0.67}/0.03/0.05 & \underline{0.27}/0.01/0.04 & \underline{0.17}/0.01/0.02 & \textbf{2.72}/0.69/0.13 & 23.14/0.25/0.12 & 2.28/0.13/0.04 & \underline{3.05}/0.04/0.03 & \underline{0.84}/0.13/0.13 & \underline{3.45}/0.13/0.19 \\
DEFDVO (ours) & ERP & \underline{0.47}/0.02/0.01 & \underline{1.32}/0.10/1.45 & \underline{1.00}/0.04/0.05 & \textbf{0.12}/0.00/0.02 & 0.27/0.09/0.05 & \underline{2.75}/0.69/0.12 & 22.20/0.25/0.12 & \textbf{1.13}/0.07/0.04 & \textbf{2.64}/0.04/0.03 & 0.96/0.27/0.16 & \textbf{3.29}/0.16/0.20 \\

\bottomrule
\end{tabular}}
\end{table*}

\subsection{Evaluation Results}
\label{sec:evaluation}
Since the CubeDVO implementation is not publicly available, comparison with CubeDVO is limited to the results reported in the original paper. All other baselines, including 360DVO, OpenVSLAM, and the pinhole methods, are evaluated using their released implementations under the same evaluation setup as DEFDVO. For 360DVO, the default variant uses native resolution ($3840{\times}1920$ on the 360DVO dataset and $2048{\times}1024$ on TartanAirV2), while the fast variant uses half resolution, matching DEFDVO ($1920{\times}960$ and $1024{\times}512$, respectively).

\noindent\textbf{Synthetic evaluation.} Tables~\ref{tab:set1} and~\ref{tab:set2} report results on the two TartanAirV2 evaluation sets. 
On Set~1 (Table~\ref{tab:set1}), DEFDVO reduces the best prior average ATE from 0.201\,m to 0.042\,m, a $4.8\times$ improvement, and is the only method with ATE below 0.12\,m on all eight sequences. The largest improvement over the ERP baselines is observed on the hard OldTownNight sequence, where DEFDVO reduces ATE from 1.260\,m with 360DVO to 0.116\,m, indicating improved robustness on the hardest inputs rather than shaving already-small errors. The pinhole baselines are accurate on the low-drift scenes but exhibit large errors on the House (hard) and OldTownNight trajectories, yielding split averages between 1.164\,m and 2.627\,m. Compared with 360DVO (fast), DEFDVO also reduces the average RPE(t) from 0.008 to 0.006\,m/frame and RPE(r) from 0.043 to 0.026\,deg/frame. While 360DVO (default) achieves the lowest RPE (0.005\,m/frame and 0.015\,deg/frame), potentially due to its higher resolution, its ATE is $4.8\times$ that of DEFDVO (0.202 vs.\ 0.042\,m). All three methods outperform the pinhole baselines in average RPE. These results suggest that DEFDVO's advantage lies in improved long-term trajectory consistency.

On Set~2 (Table~\ref{tab:set2}), DEFDVO achieves the lowest ATE on six of the twelve sequences and the best average ATE of 0.088\,m, improving over 360DVO (default) by $1.6\times$, over the best pinhole average (DPVO, 2.318\,m) by $26\times$, and over the published CubeDVO result (1.02\,m) by more than an order of magnitude. Together with Set~1, these results show that the performance gains persist across both TartanAirV2 evaluation sets. Since 360DVO excludes only its own evaluation scenes from training, most Set~2 scenes are part of its training data, potentially favoring it in this comparison.

\noindent\textbf{Real-world evaluation.}
Table~\ref{tab:360dvo_dataset} reports results on the real-world 360DVO dataset. Despite being trained exclusively on synthetic ERP imagery, DEFDVO achieves the lowest average ATE on both splits, with 3.26\,m on \textit{Easy Sequences} and 3.29\,m on \textit{Hard Sequences}. Among the pinhole methods, only DPVO successfully tracks all sequences, with \textit{Easy}/\textit{Hard} ATEs of 7.83/6.92\,m. The two 360DVO variants exhibit markedly different performance across the splits, with \textit{Easy}/\textit{Hard} ATEs of 3.60/6.06\,m for the default variant and 5.94/3.45\,m for the fast variant. In contrast, DEFDVO maintains comparable accuracy across \textit{Easy} and \textit{Hard} (3.26/3.29\,m), indicating more consistent performance across difficulty levels and strong synthetic-to-real generalization.

\begin{figure}[t]
\centering
\includegraphics[width=\columnwidth]{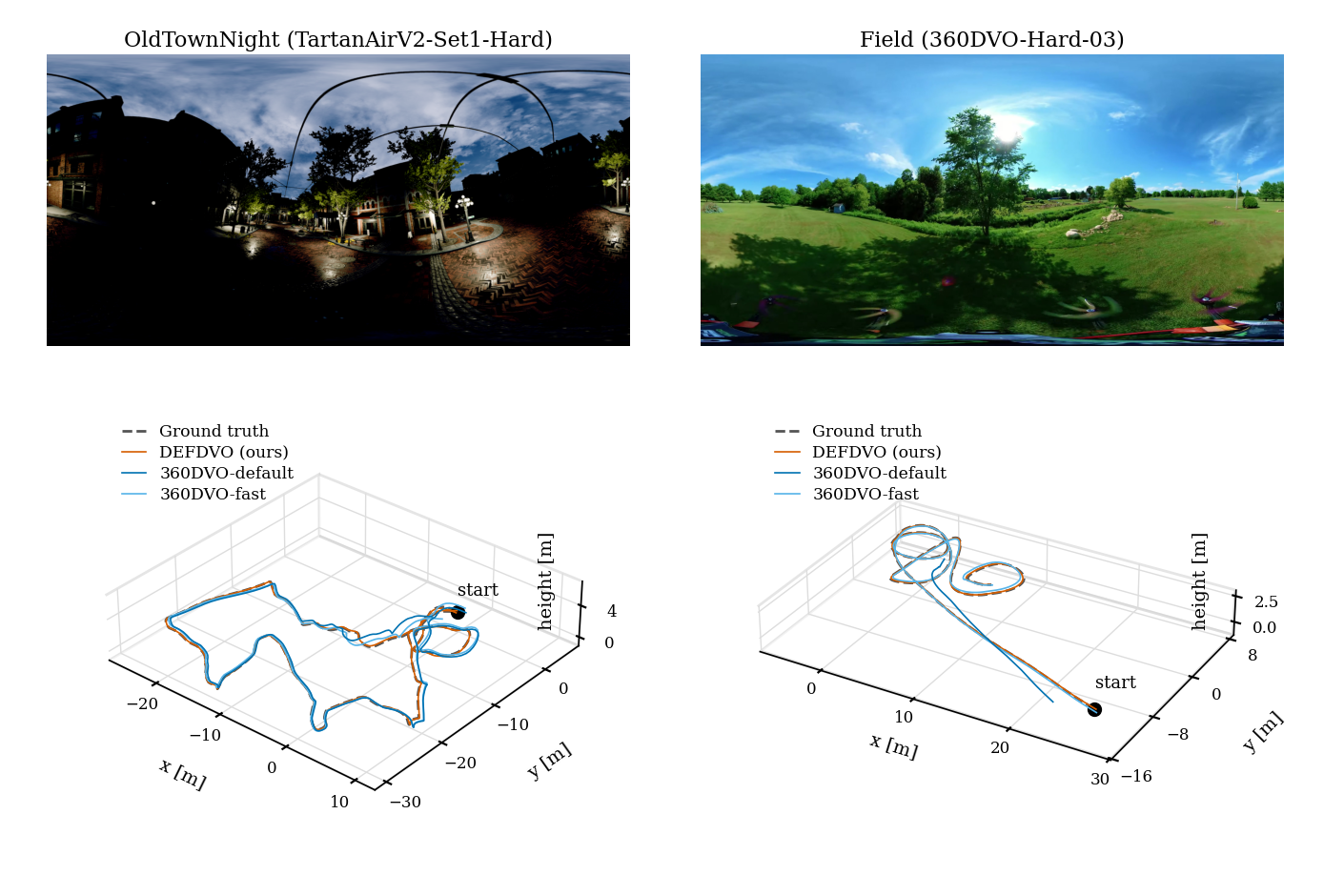}
\caption{\textbf{Qualitative trajectory comparison.} ERP frames (top) and estimated trajectories (bottom) for two evaluation sequences. All trajectories are Sim(3)-aligned to the ground truth.}
 \label{fig:traj}
\end{figure}

Fig.~\ref{fig:traj} compares the estimated trajectories of DEFDVO and both 360DVO variants on one sequence from each benchmark. DEFDVO closely follows the ground truth on both sequences, with ATEs of 0.12\,m and 0.12\,m. Among the 360DVO variants, the fast version tracks more consistently, with ATEs of 1.10\,m and 0.27\,m, while the default variant exhibits substantial drift on the Field sequence.

\noindent\textbf{Overall results.}
DEFDVO attains the lowest average ATE on every split of both benchmarks (Table~\ref{tab:summary}), reducing the average error on the TartanAirV2 evaluation sets from 0.172\,m to 0.065\,m ($2.6\times$) and on the real-world 360DVO benchmark from 4.70\,m to 3.27\,m. 

%\begin{table}[t]
%\centering
%\caption{Summary ATE comparison on the 360DVO dataset splits and the two TartanAirV2 evaluation sets. Avg.\ is the mean of the two split (set) averages. CubeDVO values are its published results; ``--'' denotes not available. Best in \textbf{bold}, second-best \underline{underlined}.}
%\label{tab:summary}
%\footnotesize
%\setlength{\tabcolsep}{2.5pt}
%\begin{tabular}{lcccccc}
%\toprule
%\multirow{2}{*}{Method} & \multicolumn{3}{c}{360DVO dataset} & \multicolumn{3}{c}{TartanAirV2} \\
%\cmidrule(lr){2-4}\cmidrule(lr){5-7}
% & Easy & Hard & Avg. & Set~1 & Set~2 & Avg. \\
%\midrule
%360DVO (def.)~\cite{guo2026_360dvo} & \underline{3.59} & 6.06 & 4.83 & 0.202 & \underline{0.142} & \underline{0.172} \\
%360DVO (fast) & 5.94 & \underline{3.45} & \underline{4.70} & \underline{0.201} & 0.421 & 0.311 \\
%CubeDVO~\cite{ando2026cubedvo} & -- & -- & -- & -- & 1.02 & -- \\
%DEFDVO (ours) & \textbf{3.26} & \textbf{3.29} & \textbf{3.27} & \textbf{0.042} & \textbf{0.088} & \textbf{0.065} \\
%\bottomrule
%\end{tabular}
%\end{table}

\begin{table}[t]
\centering
\caption{Summary ATE comparison of omnidirectional methods on the 360DVO dataset splits and the two TartanAirV2 evaluation sets. Avg.\ is the mean of the two split (set) averages. ``na'' denotes not available. ``--'' denotes tracking failure. Best ATE in \textbf{bold}, second-best \underline{underlined}.}
\label{tab:summary}
\footnotesize
\setlength{\tabcolsep}{2.5pt}
\begin{tabular}{lccccccc}
\toprule
\multirow{2}{*}{Method} & \multicolumn{3}{c}{360DVO dataset} & \multicolumn{3}{c}{TartanAirV2} & \multirow{2}{*}{FPS} \\
\cmidrule(lr){2-4}\cmidrule(lr){5-7}
 & Easy & Hard & Avg. & Set~1 & Set~2 & Avg. & \\
\midrule
OpenVSLAM~\cite{sumikura2019openvslam} & -- & -- & -- & 2.546 & 4.734 & 3.640 & 7.63 \\
CubeDVO~\cite{ando2026cubedvo} & na & na & na & na & 1.02 & na & na \\
360DVO (def.)~\cite{guo2026_360dvo} & \underline{3.60} & 6.06 & 4.83 & 0.202 & \underline{0.142} & \underline{0.172} & 5.96 \\
360DVO (fast) & 5.94 & \underline{3.45} & \underline{4.70} & \underline{0.201} & 0.421 & 0.311 & 16.62 \\
DEFDVO (ours) & \textbf{3.26} & \textbf{3.29} & \textbf{3.27} & \textbf{0.042} & \textbf{0.088} & \textbf{0.065} & 7.13 \\
\bottomrule
\end{tabular}
\end{table}

\subsection{Ablation Studies}
\label{sec:ablation}
 
\noindent\textbf{ERP adaptations.} Table~\ref{tab:ablation-component} evaluates the incremental contribution of the proposed adaptations. The \textit{Equirect.\ DPVO} baseline employs the spherical camera model with inverse range and wrapped BA residuals while preserving a standard ResNet encoder, planar-gradient patch selection, and no explicit seam-aware feature processing. Accordingly, \emph{Seam Awar.} column denotes the seam-aware correlation and circularly padded feature extraction described in Sec.~\ref{sec:seam}. All variants are trained from scratch. %double check todo

Replacing ResNet with S-ResNet (analytic spherical encoder) reduces ATE from 4.36\,m to 4.02\,m, and adding seam-awareness further improves it to 3.95\,m. Introducing the zero-initialized DCNv2 encoder reduces ATE to 3.82\,m, while geometry initialization further lowers it to 3.69\,m. This progression shows that learned deformation and the spherical sampling prior provide complementary gains. Finally, spherical-gradient patch selection yields the largest improvement, reducing ATE from 3.69\,m to 3.27\,m.

\begin{table}[t]
\centering
\caption{Ablation of the proposed adaptations on the 360DVO dataset in ATE. Each row builds on the preceding variant, with the final row corresponding to DEFDVO.}
\label{tab:ablation-component}
\footnotesize
\setlength{\tabcolsep}{3pt}
\begin{tabular}{lcccc}
\toprule
Method & \makecell{Feat.\\Extr.} & \makecell{Seam\\Awar.} & \makecell{Patch\\Sel.} & ATE  \\
\midrule
Equirect.\ DPVO    & ResNet   & \xmark & Grad.   & 4.36 \\
+ Sph.\ ResNet     & S-ResNet & \xmark & Grad.   & 4.02 \\
+ Seam Awareness   & S-ResNet & \cmark & Grad.   & 3.95 \\
+ DCNv2 (zero)     & DCNv2    & \cmark & Grad.   & 3.82 \\
+ DCNv2 (geom.)    & DCNv2    & \cmark & Grad.   & 3.69 \\
+ Patch Selection  & DCNv2    & \cmark & S-Grad. & 3.27 \\
\bottomrule
\end{tabular}
\end{table}

\begin{table}[t]
\centering
\caption{Ablation of image scale and patch selection on the 360DVO dataset in ATE. Grad. refers to gradient, while S-Grad. denotes spherical-gradient. Each training setting uses a separately trained model, while patch selection is varied at evaluation. Best per column in \textbf{bold}, second-best \underline{underlined}.}
\label{tab:setting-ablation}
\footnotesize
\setlength{\tabcolsep}{4pt}
\begin{tabular}{ccccccc}
\toprule
\multirow{2}{*}{ID} & \multirow{2}{*}{Scale}
 & \multicolumn{2}{c}{Patch Selection} & \multicolumn{3}{c}{ATE} \\
\cmidrule(lr){3-4}\cmidrule(lr){5-7}
 & & Train & Eval & Easy & Hard & Avg. \\
\midrule
1 & \multirow{4}{*}[-7pt]{0.5} & \multirow{2}{*}[-2pt]{Grad.}   & Grad.   & 3.27 & 4.10 & 3.69 \\
\cmidrule(lr){4-7}
2 &                            &                                & S-Grad. & \underline{3.21} & 4.06 & 3.63 \\
\cmidrule(lr){3-7}
3 &                            & \multirow{2}{*}[-2pt]{S-Grad.} & Grad.   & 7.59 & 3.86 & 5.72 \\
\cmidrule(lr){4-7}
4 &                            &                                & S-Grad. & 3.26 & \textbf{3.29} & \textbf{3.27} \\
\midrule
5 & \multirow{4}{*}[-7pt]{1.0} & \multirow{2}{*}[-2pt]{Grad.}   & Grad.   & 4.28 & 3.93 & 4.11 \\
\cmidrule(lr){4-7}
6 &                            &                                & S-Grad. & 3.44 & \underline{3.62} & 3.53 \\
\cmidrule(lr){3-7}
7 &                            & \multirow{2}{*}[-2pt]{S-Grad.} & Grad.   & 4.16 & 4.23 & 4.20 \\
\cmidrule(lr){4-7}
8 &                            &                                & S-Grad. & \textbf{2.76} & 3.81 & \underline{3.28} \\
\bottomrule
\end{tabular}
\end{table}

\noindent\textbf{Patch selection strategy.} Table~\ref{tab:setting-ablation} compares different patch selection strategies during training and evaluation at two image scales. Switching models trained with gradient patch selection (Grad.) to spherical-gradient selection (S-Grad.) at evaluation reduces the average ATE from 3.69 to 3.63 m at scale 0.5 and from 4.11 to 3.53 m at scale 1.0. Conversely, evaluating S-Grad.-trained models with Grad.\ increases the ATE from 3.27 to 5.72 m at scale 0.5 and from 3.28 to 4.20 m at scale 1.0. These results suggest that the learned update operator adapts to the patch distribution observed during training. Spherical-gradient selection during both training and inference gives the best average ATE at both scales. Additional inference-time ablations on patch selection, image resolution, and patch count are provided in the supplementary material.
 
\begin{figure}[t]
\centering
\includegraphics[width=0.72\columnwidth]{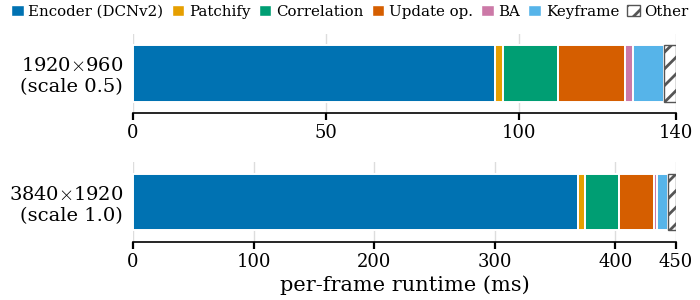}
\caption{Per-module runtime decomposition for the default and full image scales, measured on an NVIDIA RTX A5000.}
\label{fig:runtime}
\end{figure}

\noindent\textbf{Runtime.} 
Fig.~\ref{fig:runtime} decomposes the per-frame runtime of each module on an RTX A5000.
The DCNv2-based encoder dominates the runtime, increasing from 94\,ms (67\% of 140\,ms) at $1920{\times}960$ to 369\,ms (82\% of 450\,ms) at full resolution. In contrast, correlation and the update operator vary only slightly with a fixed budget of 192 patches, indicating that feature extraction limits real-time performance.

\section{Conclusion}
\label{sec:conclusion}
This work presents DEFDVO, a deep patch-based visual odometry designed for equirectangular imagery. Building on DPVO, DEFDVO introduces three key components: spherically initialized deformable feature extraction, end-to-end seam-aware estimation, and spherical-gradient patch selection. Together, these components provide distortion-aware feature sampling, consistent tracking across the ERP seam, and latitude-aware patch selection while operating directly on equirectangular imagery. 

On TartanAirV2, DEFDVO reduces the average ATE from 0.201\,m to 0.042\,m on the 360DVO evaluation split, a $4.8\times$ improvement over the best prior result. Despite training on synthetic ERP imagery, DEFDVO also achieves the lowest average ATE on both splits of the real-world 360DVO benchmark. Ablations confirm the contribution of the proposed components, with spherical-gradient patch selection yielding the largest incremental gain. Future work includes exploring more efficient deformable operators, and loop closure with global optimization toward a complete $360^\circ$ SLAM system.

\section*{Acknowledgements}
\raggedright
\small{
This work is supported by the ADVISOR ITEA 23014 project.
}
{
    \small
    \bibliographystyle{ieeenat_fullname}
    \bibliography{main}
}

\clearpage
\appendix

\maketitlesupplementary

% macros shared with sec/3_method.tex (\providecommand: no-op if already defined)
\providecommand{\wrap}{\operatorname{wrap}}
\providecommand{\atantwo}{\operatorname{atan2}}
\providecommand{\SE}{\mathrm{SE}(3)}
\providecommand{\Sim}{\mathrm{Sim}(3)}

% lettered supplementary sections; floats/equations continue the main paper
\renewcommand{\thesection}{\Alph{section}}
\renewcommand{\thesubsection}{\Alph{section}.\arabic{subsection}}

\renewcommand{\wrap}{\operatorname{wrap}}
\renewcommand{\atantwo}{\operatorname{atan2}}
\renewcommand{\SE}{\mathrm{SE}(3)}
\renewcommand{\Sim}{\mathrm{Sim}(3)}
\newcommand{\skewm}[1]{[#1]_{\times}}

\section{Bundle-Adjustment Jacobians}
\label{sec:supp_jacobians}

Each Gauss--Newton iteration of the bundle adjustment (Sec.~3.4 of the
main paper) linearizes the seam-aware residual $\mathbf{r}_{kj}$ of
Eq.~(5) with respect to three quantities: the host pose $\mathbf{T}_i$,
the target pose $\mathbf{T}_j$, and the patch inverse range $d_k$. This
section derives the three Jacobians. We use the notation of the main
paper: poses are world-to-camera transformations in $\SE$, the relative
motion is
$\mathbf{G}_{ij}=\mathbf{T}_j\mathbf{T}_i^{-1}=(\mathbf{R}_{ij},\mathbf{t}_{ij})$,
and a patch $k$ with unit bearing $\mathbf{n}_k$ and inverse range $d_k$
maps to the scale-free point
$\mathbf{x}_{kj}=\mathbf{R}_{ij}\mathbf{n}_k+d_k\mathbf{t}_{ij}$ of
Eq.~(2), which is reprojected as
$\mathbf{u}'_{kj}=\Pi(\mathbf{x}_{kj})=(u'_{kj},v'_{kj})^{\top}$ by
Eq.~(3).

The residual depends on all three quantities \emph{only through} the
point $\mathbf{x}_{kj}$. By the chain rule, each Jacobian therefore
factors into the projection Jacobian
$\mathbf{J}_\Pi=\partial(u',v')/\partial\mathbf{x}$, already given in
Eq.~(7) of the main paper, times a derivative of the point,
\[
\frac{\partial\mathbf{u}'_{kj}}{\partial\theta}
=\mathbf{J}_\Pi\,\frac{\partial\mathbf{x}_{kj}}{\partial\theta},
\]
so only the point derivatives
$\partial\mathbf{x}_{kj}/\partial\theta$ remain to be computed. All
Jacobians are evaluated at the patch center, which carries the patch's
shared inverse range.

\noindent\textbf{Pose parameterization.}
Bundle adjustment does not differentiate with respect to the entries of
a pose matrix. Instead, a pose update is parameterized by a
six-dimensional vector applied through the left-multiplicative
retraction used by DROID-SLAM and DPVO,
\begin{equation}
\mathbf{T}\leftarrow\exp(\boldsymbol{\delta\xi}^{\wedge})\,\mathbf{T},
\qquad
\boldsymbol{\delta\xi}
=\begin{pmatrix}\boldsymbol{\delta\tau}\\
\boldsymbol{\delta\omega}\end{pmatrix}\in\mathbb{R}^{6},
\label{eq:supp_retraction}
\end{equation}
with translational component $\boldsymbol{\delta\tau}$ and rotational
component $\boldsymbol{\delta\omega}$, where
$\boldsymbol{\delta\xi}^{\wedge}
=\begin{psmallmatrix}
\skewm{\boldsymbol{\delta\omega}} & \boldsymbol{\delta\tau}\\
\mathbf{0}^{\top} & 0
\end{psmallmatrix}$
and $\skewm{\cdot}$ is the $3{\times}3$ skew-symmetric matrix of a
3-vector. Intuitively, $\exp(\boldsymbol{\delta\xi}^{\wedge})$ composes
a small rigid motion with the current pose: to first order,
$\exp(\boldsymbol{\delta\xi}^{\wedge})
=\mathbf{I}_4+\boldsymbol{\delta\xi}^{\wedge}$, and a point
$\mathbf{q}$ expressed in the camera frame moves to
$\mathbf{q}+\boldsymbol{\delta\tau}
+\boldsymbol{\delta\omega}\times\mathbf{q}$. All Jacobians below are
derivatives with respect to $\boldsymbol{\delta\xi}$, evaluated at
$\boldsymbol{\delta\xi}=\mathbf{0}$.

\noindent\textbf{Target pose.}
Perturbing the target pose multiplies the relative motion from the
left,
$\mathbf{G}_{ij}\leftarrow\exp(\boldsymbol{\delta\xi}_j^{\wedge})\mathbf{G}_{ij}$:
the small motion acts \emph{in the target frame} after the point has
been transformed. The point is therefore displaced directly, to first
order, by
$d_k\boldsymbol{\delta\tau}_j
+\boldsymbol{\delta\omega}_j\times\mathbf{x}_{kj}$.
The translation is weighted by $d_k$ because $\mathbf{x}_{kj}$ is
scale-free. The metric point $\mathbf{x}_{kj}/d_k$ moves by the
usual $\boldsymbol{\delta\tau}_j$, so distant patches are
insensitive to camera translation. Writing
$\boldsymbol{\delta\omega}_j\times\mathbf{x}_{kj}
=-\skewm{\mathbf{x}_{kj}}\boldsymbol{\delta\omega}_j$ and collecting
the translational and rotational blocks gives
\begin{equation}
\frac{\partial\mathbf{x}_{kj}}{\partial\boldsymbol{\delta\xi}_j}
=\big[\,d_k\mathbf{I}_3 \;\big|\; -\skewm{\mathbf{x}_{kj}}\,\big]
\in\mathbb{R}^{3\times6}.
\label{eq:supp_jj}
\end{equation}

\noindent\textbf{Host pose.}
Perturbing the host pose enters through its inverse,
$\mathbf{G}_{ij}\leftarrow\mathbf{G}_{ij}\exp(-\boldsymbol{\delta\xi}_i^{\wedge})$:
the small motion acts \emph{in the host frame}, before the point is
transformed. To first order it moves the hosted point,
$\mathbf{n}_k\to\mathbf{n}_k-d_k\boldsymbol{\delta\tau}_i
-\boldsymbol{\delta\omega}_i\times\mathbf{n}_k$ with $d_k$ unchanged,
and the rotation $\mathbf{R}_{ij}$ maps this displacement into the
target frame. With
$-\mathbf{R}_{ij}(\boldsymbol{\delta\omega}_i\times\mathbf{n}_k)
=\mathbf{R}_{ij}\skewm{\mathbf{n}_k}\boldsymbol{\delta\omega}_i$,
\begin{equation}
\frac{\partial\mathbf{x}_{kj}}{\partial\boldsymbol{\delta\xi}_i}
=\big[\,-d_k\mathbf{R}_{ij} \;\big|\; \mathbf{R}_{ij}\skewm{\mathbf{n}_k}\,\big].
\label{eq:supp_ji}
\end{equation}
The mirrored structure of Eqs.~\eqref{eq:supp_jj} and
\eqref{eq:supp_ji} reflects the same motion applied on the opposite
side of $\mathbf{G}_{ij}$. The two Jacobians are related through the
$\SE$ adjoint of the relative motion,
\begin{equation}
\frac{\partial\mathbf{x}_{kj}}{\partial\boldsymbol{\delta\xi}_i}
=-\,\frac{\partial\mathbf{x}_{kj}}{\partial\boldsymbol{\delta\xi}_j}\,
\mathrm{Ad}_{\mathbf{G}_{ij}},
\quad
\mathrm{Ad}_{\mathbf{G}_{ij}}
=\begin{psmallmatrix}
\mathbf{R}_{ij} & \skewm{\mathbf{t}_{ij}}\mathbf{R}_{ij}\\
\mathbf{0} & \mathbf{R}_{ij}
\end{psmallmatrix},
\label{eq:supp_adj}
\end{equation}
which provides an independent consistency check on both derivations.

\noindent\textbf{Inverse range.}
$\mathbf{x}_{kj}$ is affine in $d_k$, so
\begin{equation}
\frac{\partial\mathbf{x}_{kj}}{\partial d_k}=\mathbf{t}_{ij}.
\label{eq:supp_jd}
\end{equation}
Increasing the inverse range moves the point along the relative
translation, the direction of parallax. As $d_k\to0$ the translation
blocks of Eqs.~\eqref{eq:supp_jj} and \eqref{eq:supp_ji} vanish and
$\mathbf{x}_{kj}\to\mathbf{R}_{ij}\mathbf{n}_k$ remains a unit vector,
so distant patches degrade gracefully to pure rotation constraints
(Sec.~3.2 of the main paper).

\noindent\textbf{Residual sign and masking.}
The $\wrap$ operator of Eq.~(4) of the main paper has a unit derivative
almost everywhere, and the predicted target $\mathbf{u}^\star_{kj}$ is
constant, so the residual Jacobians are the negated reprojection
Jacobians,
$\partial\mathbf{r}_{kj}/\partial\theta
=-\,\mathbf{J}_\Pi\,\partial\mathbf{x}_{kj}/\partial\theta$. Each
damped Gauss--Newton iteration assembles the normal equations from
these blocks with the confidence weights $\mathbf{w}_{kj}$ of Eq.~(6)
and proceeds as in Sec.~3.4 of the main paper. Measurements are masked
when the transformed point passes closer than a minimum range to the
target camera, leaves the padded vertical image bounds, or approaches
an ERP pole ($\rho=\sqrt{X^2+Z^2}\to0$), where $\mathbf{J}_\Pi$
diverges; the latitude cap bounds host latitudes only, so the target
elevation $\phi'$ is guarded at run time.

\noindent\textbf{Numerical verification.}
The closed-form Jacobians of
Eqs.~\eqref{eq:supp_jj}--\eqref{eq:supp_jd}, combined with
$\mathbf{J}_\Pi$, match automatic differentiation of the projection
through the $\SE$ exponential map to double-precision accuracy (maximum
absolute deviation $2.3\times10^{-13}$ over 100 random
pose/bearing/range configurations), and the identity in
Eq.~\eqref{eq:supp_adj} holds to machine precision.

\section{Additional Implementation Details}
\label{sec:supp_impl}

\noindent\textbf{Training.}
Table~\ref{tab:train_hparams} lists the training configuration. The
recurrent patch graph is initialized on the first eight frames. After the eighth update iteration, one frame is appended per iteration until all 15 frames are included, and the remaining iterations refine the full graph. During the first 1k steps, only the patch inverse ranges are optimized, with poses fixed to ground truth. The pose loss is applied from the second update iteration onward, after $\Sim$ alignment with the ground truth trajectory. Planar crop and scale augmentation are disabled, and processing scales are restricted so that $f_x=W'/(2\pi)$ holds exactly at every scale (Sec.~4.3 of the main paper). 

\begin{table}[t]
\centering
\caption{Training hyperparameters.}
\label{tab:train_hparams}
\footnotesize
\setlength{\tabcolsep}{6pt}
\begin{tabular}{ll}
\toprule
Optimizer & AdamW \\
Learning rate & $8\times10^{-5}$, one-cycle (linear, 1\% warmup) \\
Weight decay & $10^{-6}$ \\
Gradient clipping & 10 (global norm) \\
Training steps & 120k \\
Frames per sequence & 15 \\
Patches per frame & 96 (spherical-gradient) \\
Update iterations & 18 per sequence \\
BA iterations & 2 per update \\
GN damping & $\lambda=10^{-4}$, $+10$ on the pose diagonal \\
Loss weights & 0.1 flow, 10 pose \\
Image scale & 0.5 ($1024{\times}512$) \\
Structure-only steps & first 1k \\
Frame-pair flow range & 10--75\,px \\
\bottomrule
\end{tabular}
\end{table}

\noindent\textbf{Inference.}
DEFDVO maintains DPVO's default scheduling: a sliding optimization window of 10 keyframes, a patch lifetime of 13 frames, edge removal after 22 frames, keyframing by mean flow magnitude below 15\,px (computed with the seam-aware wrap), a damped constant-velocity motion model (damping 0.5), and initialization after eight frames with sufficient motion accumulated, with 12 update iterations at initialization and 12 more at termination. 
% It differs in the patch budget and selection: 192 patches per frame with spherical-gradient selection ($3\times$ oversampling) and in the circular azimuth indexing of the $7{\times}7$ two-level correlation lookup (Sec.~3.3 of the main paper). 
DEFDVO differs in its patch budget and selection strategy, using 192 patches per frame with spherical-gradient selection, as well as in its circular azimuth indexing for the $7{\times}7$ two-level correlation lookup (Sec.~3.3 of the main paper). The correlation and update operator run under automatic mixed precision, while patch extraction does not.

\section{Additional Inference-Time Ablations}
\label{sec:supp_ablation}

This section complements the ablation studies of Sec.~4.5 in the main
paper with inference-time ablations on the real-world 360DVO dataset. All rows of Table~\ref{tab:hyperparams} use the same default-trained model and vary only in the inference configuration. \#D denotes the default setting ($1920{\times}960$ resolution, $N{=}192$ patches,
spherical-gradient selection). ATE is reported in meters after $\Sim$
alignment, following the protocol of the main paper, and FPS is measured
on an NVIDIA RTX A5000.

\begin{table}[t]
\centering
\caption{Ablation of DEFDVO inference-time hyperparameters on the 360DVO
dataset in ATE. All rows use the same default-trained model. \#D denotes the default setting used in all other experiments, and \#R, \#P, and \#S vary image resolution, patch count ($N$), and patch selection, respectively.}
\label{tab:hyperparams}
\footnotesize
\setlength{\tabcolsep}{2pt}
\begin{tabular}{lccccccc}
\toprule
\multirow{2}{*}{ID} & \multicolumn{3}{c}{Settings} & \multicolumn{3}{c}{ATE} & \multirow{2}{*}{FPS} \\
\cmidrule(lr){2-4}\cmidrule(lr){5-7}
 & Resolution & $N$ & Sel. & Easy & Hard & Avg. & \\
\midrule
\#D  & 1920$\times$960  & 192 & S-Grad. & 3.26 & 3.29 & 3.27 & 7.13 \\
\midrule
\#R0 & 3840$\times$1920 & 192 & S-Grad. & 3.10 & 4.85 & 3.98 & 2.23 \\
\#R1 & 960$\times$480   & 192 & S-Grad. & 4.32 & 3.50 & 3.91 & 18.04 \\
\midrule
\#P0 & 1920$\times$960  & 384 & S-Grad. & 3.40 & 3.30 & 3.35 & 5.64 \\
\#P1 & 1920$\times$960  & 96  & S-Grad. & 3.30 & 3.72 & 3.51 & 8.08 \\
\#P2 & 1920$\times$960  & 48  & S-Grad. & 3.16 & 4.25 & 3.71 & 8.57 \\
\midrule
\#S0 & 1920$\times$960  & 192 & Grad.   & 7.59 & 3.86 & 5.72 & 6.89 \\
\#S1 & 1920$\times$960  & 192 & Random  & 8.46 & 17.29 & 12.88 & 6.83 \\
\bottomrule
\end{tabular}
\end{table}

\noindent\textbf{Resolution and patch count.}
Increasing to full resolution (\#R0) improves the easy split but
substantially degrades the hard split (4.85\,m), while reducing
throughput to roughly one third of the default. Halving the resolution
(\#R1) also reduces accuracy. Varying the patch count shows that
$N{=}192$ provides the best accuracy-runtime trade-off: smaller budgets
(\#P1, \#P2) trade accuracy for throughput, and doubling to $N{=}384$
(\#P0) does not improve accuracy.

\noindent\textbf{Patch selection.}
Rows \#S0 and \#S1 replace spherical-gradient selection at inference
while keeping the default resolution and patch count. Planar gradient
selection (\#S0) increases the average ATE from 3.27\,m to 5.72\,m,
consistent with the cross-ablation in Table~6 of the main paper, and
random selection (\#S1) performs far worse, reaching 12.88\,m,
confirming that latitude-aware selection is required at inference.

\section{Latitude Cap in Patch Selection}
\label{sec:supp_cap}

Equation~(9) of the main paper samples candidate latitudes only within
$|\phi|\le75^\circ$. This section shows what this cap excludes and how
it affects the selected patches.

\begin{table}[t]
\centering
\caption{Regions excluded by the polar cap $\phi_{\max}$ of Eq.~(9) in
the main paper: shares of the sphere and of the ERP image rows with
$|\phi|>\phi_{\max}$ (both poles together), and the horizontal
reprojection sensitivity $1/\cos\phi_{\max}$ relative to the equator
(Sec.~3.4 of the main paper).}
\label{tab:cap_geometry}
\footnotesize
\setlength{\tabcolsep}{6pt}
\begin{tabular}{lccc}
\toprule
$\phi_{\max}$ & Sphere & ERP rows & $1/\cos\phi_{\max}$ \\
\midrule
$65^\circ$           & 9.4\% & 27.8\% & 2.37  \\
$75^\circ$ (default) & 3.4\% & 16.7\% & 3.86  \\
$85^\circ$           & 0.4\% & 5.6\%  & 11.47 \\
\bottomrule
\end{tabular}
\end{table}

\begin{table}[t]
\centering
\caption{Shares of candidates and of selected patch centers with
$|\phi|$ above the given latitude, for random (\#S1), planar gradient
(\#S0), and spherical-gradient selection without a cap and with the
default cap (\#D); IDs as in Table~\ref{tab:hyperparams}. Measured at
the default resolution on every frame of the 20 360DVO sequences
(20{,}008 frames, $1920{\times}960$) and of the 19 TartanAirV2
evaluation trajectories (12{,}179 frames, $1024{\times}512$), with one
draw per frame as in tracking, i.e., 3.84\,M and 2.34\,M selected
patches per row. Random selection does not rank candidates, so its
candidates are its patches.}
\label{tab:cap_selection}
\footnotesize
\setlength{\tabcolsep}{3.5pt}
\begin{tabular}{llcccc}
\toprule
\multirow{2}{*}{Sel.} & \multirow{2}{*}{Cap} & Candidates
 & \multicolumn{3}{c}{Selected patches} \\
\cmidrule(lr){3-3}\cmidrule(lr){4-6}
 & & $>75^\circ$ & $>65^\circ$ & $>75^\circ$ & $>85^\circ$ \\
\midrule
\multicolumn{6}{l}{\emph{360DVO}} \\
Random  & none       & 16.0\% & 26.9\%    & 16.0\%  & 5.0\%  \\
Grad.   & none       & 16.0\% & 3.6\%     & 0.91\%  & 0.06\% \\
S-Grad. & none       & 3.4\%  & $<$0.01\% & 0       & 0      \\
S-Grad. & $75^\circ$ & 0      & $<$0.01\% & 0       & 0      \\
\midrule
\multicolumn{6}{l}{\emph{TartanAirV2}} \\
Random  & none       & 15.9\% & 27.0\%    & 15.9\%  & 4.8\%  \\
Grad.   & none       & 15.9\% & 3.8\%     & 0.58\%  & 0.19\% \\
S-Grad. & none       & 3.4\%  & 0.04\%    & 0.002\% & 0      \\
S-Grad. & $75^\circ$ & 0      & 0.04\%    & 0       & 0      \\
\bottomrule
\end{tabular}
\end{table}

\noindent\textbf{What the cap excludes.}
An ERP image gives every latitude band of equal width the same number
of rows, but the area of such a band shrinks with $\cos\phi$ toward the
poles. The regions beyond $75^\circ$ therefore hold 16.7\% of the rows
but only 3.4\% of the sphere, and the horizontal reprojection
sensitivity there is at least 3.86 times its equatorial value
(Table~\ref{tab:cap_geometry}).

\noindent\textbf{Effect on the selected patches.}
Random (\#S1) and planar gradient (\#S0) selection sample ERP rows
uniformly, so about 16\% of their candidates lie beyond $75^\circ$
(Table~\ref{tab:cap_selection}). Random selection keeps all of them,
and planar gradient ranking still keeps 0.6--0.9\% of its patches
there. Spherical-gradient selection without the cap draws only 3.4\% of
its candidates there, their share of the sphere, and the $\cos\phi$
weight of Eq.~(10) removes almost all of them: no selected patch lies
beyond $75^\circ$ on the 360DVO dataset, and only 48 of 2.34\,M (0.002\%) do on TartanAirV2. With or without the cap, more than 99.8\% of the selected patches lie within $\pm60^\circ$. Consistently, the average ATE on 360DVO dataset does not change significantly with caps of $65^\circ$, $75^\circ$, $85^\circ$, or none.

\noindent\textbf{Role of the cap.}
The cap is therefore a safeguard rather than a tuned threshold: for any
image content, it bounds the horizontal sensitivity at every patch
center by about $1/\cos75^\circ\approx3.9$.

\section{Baseline Details}
\label{sec:supp_baselines}

All baselines except CubeDVO run from public implementations with their authors' released weights and default parameters, unless stated otherwise below. CubeDVO has no public implementation, so Tables~2 and~4 of the main paper report the results published by its authors.

\noindent\textbf{Pinhole baselines.}
DROID-SLAM, DPVO, MASt3R-SLAM, and VGGT-SLAM receive $640{\times}640$ perspective images with a $90^\circ$ field of view ($f_x{=}f_y{=}320$). On TartanAirV2, these are the images of the front camera of the six-camera rig whose views form the ERP frames. On the 360DVO dataset, one such view is rendered from each ERP frame through the panorama's optical center, so the ground-truth positions apply unchanged, and it faces the same direction on the ERP equator in every sequence. DPVO runs without loop closure, DROID-SLAM with its final global bundle adjustment, and MASt3R-SLAM and VGGT-SLAM with their loop closure. MASt3R-SLAM uses the known intrinsics and processes every frame, cropped to its network input of $512{\times}384$. VGGT-SLAM estimates its own intrinsics and returns poses only for its keyframes, so its ATE and RPE are computed on its keyframes, with RPE between consecutive keyframes. 

\noindent\textbf{OpenVSLAM.}
OpenVSLAM runs from a public fork of its original code, with the fork's GPS extension unused. It uses its equirectangular camera model at the native ERP resolution, $2048{\times}1024$ on TartanAirV2 and $3840{\times}1920$ on the 360DVO dataset, with 2000 and 4000 ORB features per frame, respectively; the top and bottom 10\% of the image rows are masked, and loop closure is enabled. It runs on the CPU.

\noindent\textbf{360DVO.}
Both variants use the released checkpoint and configurations. 

\end{document}